%% file: main.tex
\documentclass[10pt,twocolumn,letterpaper]{article}

\usepackage{iccv}
\usepackage{times}
\usepackage{epsfig}
\usepackage{graphicx}
\usepackage{amsmath}
\usepackage{amssymb}
\usepackage{pifont}
\usepackage{caption}
\usepackage{makecell}
\usepackage{subcaption}
\usepackage{etoolbox}
\usepackage[dvipsnames]{xcolor}
\usepackage{adjustbox}
\usepackage{comment}
\usepackage{multirow}
\usepackage[most]{tcolorbox}
\usepackage{cuted}
\usepackage{lipsum}

\usepackage{booktabs}

\usepackage[pagebackref=true,breaklinks=true,letterpaper=true,colorlinks,bookmarks=false]{hyperref}

\iccvfinalcopy %

\def\iccvPaperID{6900} %

\ificcvfinal\fi

\begin{document}

\title{Encyclopedic VQA:\\{\large Visual questions about detailed properties of fine-grained categories}}

\author{\stepcounter{footnote} %
Thomas Mensink\thanks{Equal contribution. $^\ddagger$Work done during internship at Google.\\\indent $^*$Google Research} $^{,*}$\\
{\tt\small mensink@google.com}
\and
Jasper Uijlings$^{\dag,*}$\\
{\tt\small jrru@google.com}
\and
Lluis Castrejon$^*$\\
{\tt\small lluisc@google.com}
\and
Arushi Goel$^\ddagger$\\
{\tt\small goel.arushi@gmail.com}
\and
Felipe Cadar$^\ddagger$\\
{\tt\small cadar@dcc.ufmg.br}
\and
Howard Zhou$^*$\\
{\tt\small howardzhou@google.com}
\and
Fei Sha$^*$\\
{\tt\small fsha@google.com}
\and
André Araujo$^*$\\
{\tt\small andrearaujo@google.com}
\and
Vittorio Ferrari$^*$\\
{\tt\small vittoferrari@google.com}
}

\maketitle
\ificcvfinal\thispagestyle{empty}\fi

\input{macros}

\input{0_abstract.tex}
\input{01_introduction.tex}

\input{02_design_principles}

\input{03_related_work.tex}
\input{04_dataset.tex}

\input{05_experiments.tex}
\input{06_conclusions.tex}

{\small
\bibliographystyle{ieee_fullname}
\bibliography{shortstrings,loco,loco_extra}
}

\clearpage
\appendix
\input{A1_user_study}

\input{A2_clip_retrieval}
\input{A3_qualitative}
\input{A4_dataset_statistics}

\input{A5_prompts}

\end{document}

%% file: macros.tex
\providetoggle{showcomments}
\settoggle{showcomments}{false} 								%

\iftoggle{showcomments}{%
    \newcommand{\resolved}[3][]{\ifstrequal{#1}{resolved}{\textcolor{blue}{RESOLVED:}~\textbf{{\MakeUppercase #2:}}~{#3}}{\textbf{\MakeUppercase #2:}~#3}}
    \newcommand{\andre}[2][]{\textcolor{ForestGreen}{\resolved[#1]{andre}{#2}}}
    \newcommand{\vitto}[2][]{\textcolor{red}{\resolved[#1]{vitto}{#2}}}
    \newcommand{\jasper}[2][]{\textcolor{violet}{\resolved[#1]{jasper}{#2}}}
    \newcommand{\jrru}[2][]{\textcolor{violet}{\resolved[#1]{jasper}{#2}}}
    \newcommand{\tm}[2][]{\textcolor{magenta}{\resolved[#1]{TM}{#2}}}
    \newcommand{\lluis}[2][]{\textcolor{RedOrange}{\resolved[#1]{lluis}{#2}}}
    \newcommand{\fei}[2][]{\textcolor{MidnightBlue}{\resolved[#1]{fei}{#2}}}
    \newcommand{\arushi}[2][]{\textcolor{YellowGreen}{\resolved[#1]{arushi}{#2}}}
    \newcommand{\changed}[1]{\textcolor{blue}{#1}}
    \newcommand{\todo}[1]{\textcolor{blue}{\textbf{TODO:} #1}}
}{%
    \newcommand{\changed}[1]{#1}
    \newcommand{\todo}[1]{}
    \newcommand{\andre}[2][]{}
    \newcommand{\vitto}[2][]{}
    \newcommand{\jasper}[2][]{}
    \newcommand{\jrru}[2][]{}
    \newcommand{\tm}[2][]{}
    \newcommand{\lluis}[2][]{}
    \newcommand{\fei}[2][]{}
    \newcommand{\arushi}[2][]{}
    \newcommand{\att}[1]{#1}
}

\newcolumntype{H}{>{\setbox0=\hbox\bgroup}c<{\egroup}@{}}

\newcommand{\kivqa}{\textit{Encyclopedic}-VQA\xspace}
\newcommand{\evqa}{\kivqa}
\newcommand{\qapair}{$Q\!+\!A$\xspace}
\newcommand{\para}[1]{\par\noindent\textbf{#1}}

\newcommand{\cmark}{\ding{51}}%
\newcommand{\xmark}{\ding{55}}%

\newtcolorbox[auto counter]{promptfloat}[2][]{%
    float=tp,%
    blend before title=dash hang,%
    title={\textbf{Prompt~\thetcbcounter:} #2},%
    colback=yellow!10!white,%
    colframe=red!75!black,%
    #1}

%% file: 0_abstract.tex
\begin{abstract}

We propose \kivqa, a large scale visual question answering (VQA) dataset
featuring visual questions about detailed properties of fine-grained categories and instances.
It contains 221k unique question+answer pairs each matched with (up to) 5 images, resulting in a total of 1M VQA samples.
Moreover, our dataset comes with a controlled knowledge base derived from Wikipedia, marking the evidence to support each answer.
Empirically, we show that our dataset poses a hard challenge for large vision+language models as they perform poorly on our dataset: PaLI~\cite{chen22arxiv} is state-of-the-art on OK-VQA~\cite{marino19cvpr}, yet it only achieves $13.0\%$ accuracy on our dataset. 
Moreover, we experimentally show that progress on answering our encyclopedic questions can be achieved by augmenting large models with a mechanism that retrieves relevant information from the knowledge base.
An oracle experiment with perfect retrieval achieves $87.0\%$ accuracy on the single-hop portion of our dataset, and an automatic retrieval-augmented prototype yields $48.8\%$.
We believe that our dataset enables future research on retrieval-augmented vision+language models.
It is available at \url{https://github.com/google-research/google-research/tree/master/encyclopedic_vqa}.

\end{abstract}

%% file: 01_introduction.tex
\section{Introduction}

\begin{figure}[t]
    \centering
    \vspace{-.3cm}
    \includegraphics[width=\linewidth]{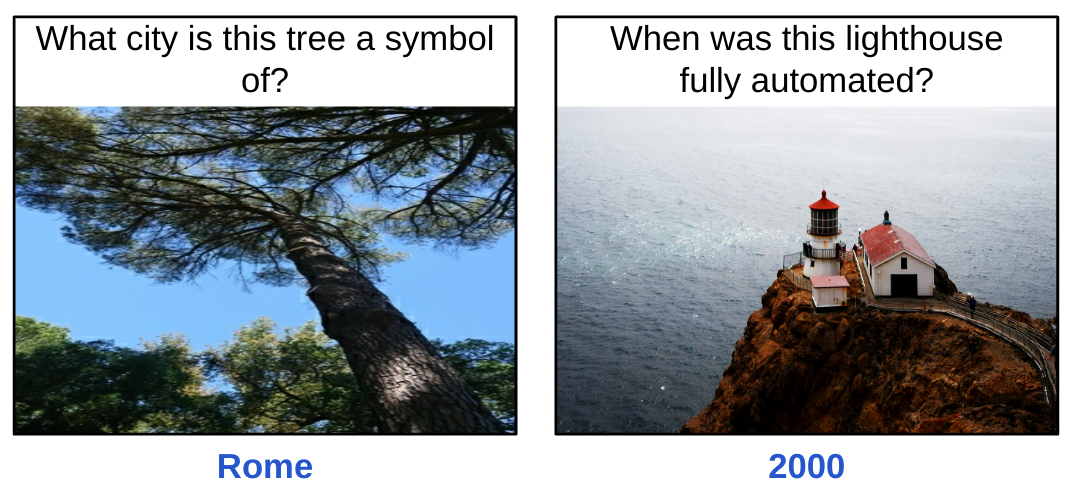}
    \caption{\small
    \textbf{One of the two answers above is wrong, do you know which one?}
    Encyclopedic questions about detailed properties of fine-grained entities are difficult. Not only for humans, but also for large VLMs.   
    PaLI~\cite{chen22arxiv} fails to answer both questions correctly.
    }
    \label{fig:wrong_pali_predictions}
    \vspace{-.2cm}
\end{figure}

Recently, large Vision+Language models (VLMs) have demonstrated impressive performance on Visual Question Answering (VQA) benchmarks~\cite{alayrac22neurips,chen22arxiv,hao22arxiv,yang22aaai}. However, Fig.~\ref{fig:wrong_pali_predictions} shows two typical examples where such models fail. Answering these questions correctly requires knowledge of detailed properties (\ie symbol of which city, year of automation) of fine-grained categories (`Pinus Pinea') or instances (`Point Reyes Lighthouse').
We hypothesize that this type of \emph{encyclopedic} knowledge is hard for VLMs to properly encode in its model parameters because such long-tail information occurs rarely in its training data (\ie the web).
Additionally, VLMs produce such incorrect answers generally with high confidence, while these answers are hard for users to verify since the model does not provide any explanations.
The answer to Fig.~\ref{fig:wrong_pali_predictions} (left) is correct; (right) should be 1975. PaLI~\cite{chen22arxiv} predicts both incorrectly.

\begin{figure*}[t]
    \centering
    \vspace{-.8cm}
    \includegraphics[width=\linewidth]{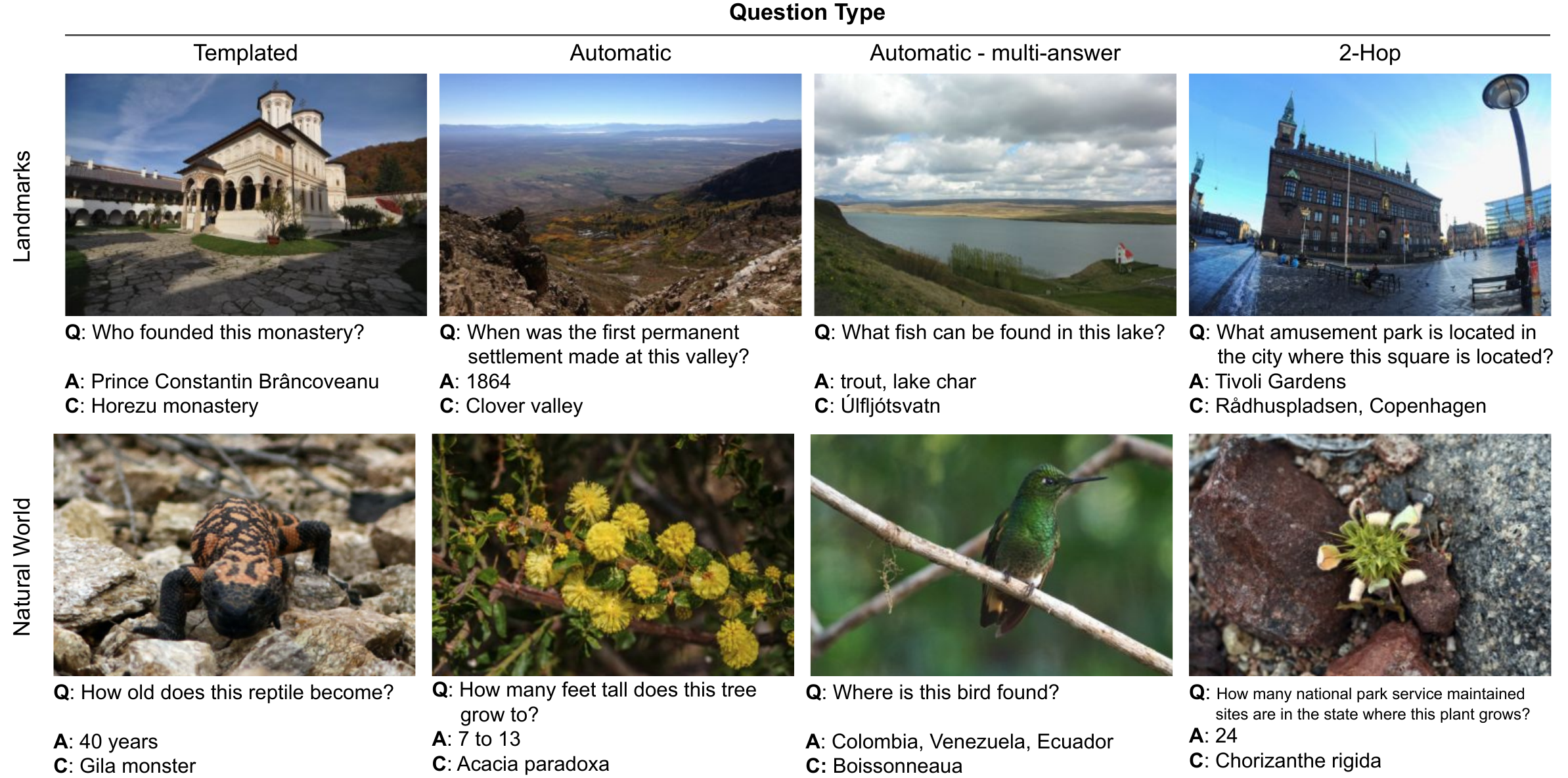}
    \caption{\small
    \textbf{Example VQA annotations for different question types.} Each example consists of an image $I$, a question $Q$ and the answer $A$. We also show the category $C$ of the subject of the question.
    As attribution, we provide a section within the Wikipedia page of $C$ which supports the answer.
    Our \kivqa dataset has a total of 1M $(I,Q,A)$ triplets.
    }
    \label{fig:our_dataset}
\end{figure*}

\begin{figure*}[ht]
    \centering
    \vspace{-.3cm}
    \includegraphics[width=0.95\linewidth]{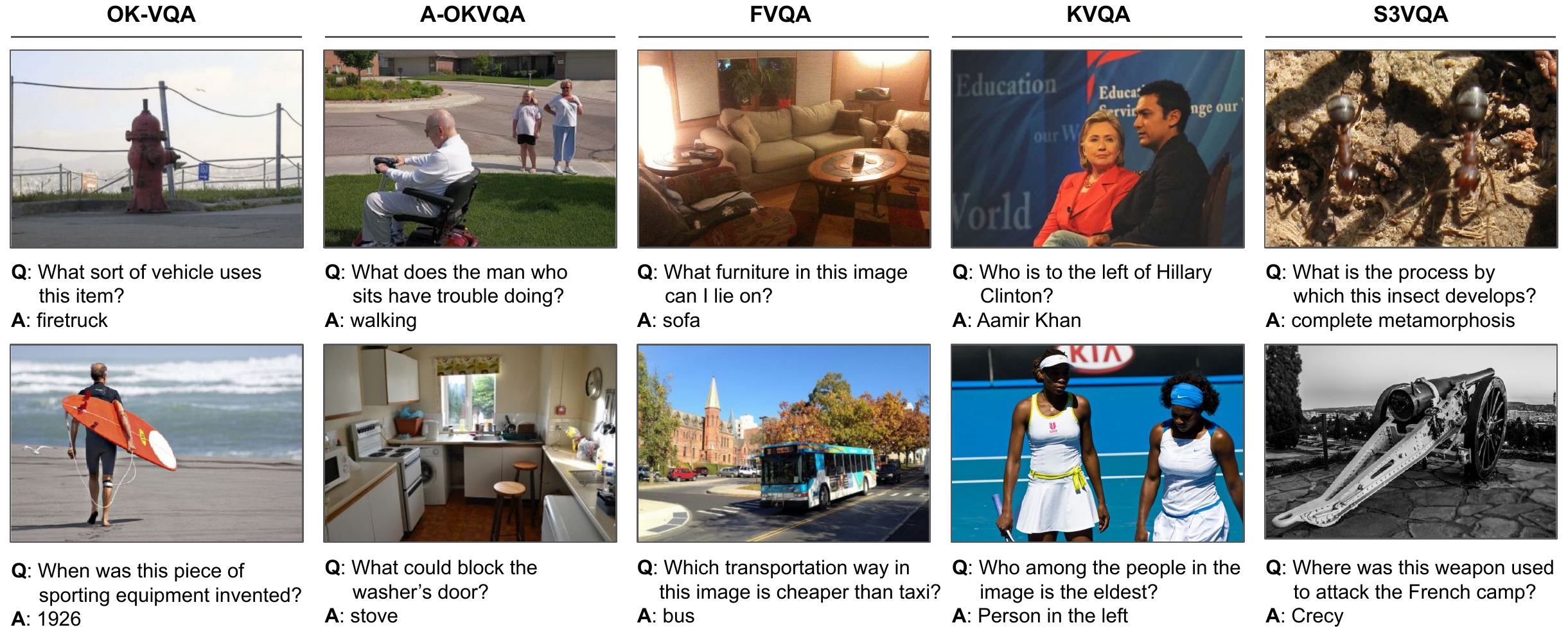}
    \caption{\small \textbf{Typical examples of VQA datasets which require knowledge beyond the image.}
    }
    \label{fig:other_datasets}
    \vspace{-.2cm}
\end{figure*}

Both problems can be addressed by retrieval-augmented models, which base their predictions on knowledge retrieved from a database.
These models recently gained popularity in NLP (\eg~\cite{borgeaud22arxiv,guu20icml,khandelwal20iclr,lewis20neurips}), and some early multi-modal models also exist~\cite{gui22naacl,hu23cvpr,marino21cvpr,vickers21acl}.
Retrieval-augmented models are well-suited for encyclopedic knowledge since retrieving the correct entry from a knowledge base greatly facilitates constructing the right answer.
Furthermore, the retrieved piece of knowledge provides \emph{attribution} to the answer by design. 
This increases model interpretability and therefore human trust, has applications to fairness and helps diagnosing and debugging model errors.

To drive progress on handling encyclopedic knowledge and attribution in VQA we need a suitable dataset.
The popular OK-VQA~\cite{marino19cvpr} and A-OKVQA~\cite{schwenk22eccv} datasets focus on questions requiring knowledge outside the query image.
However, the majority of questions require commonsense knowledge (\eg Fig.~\ref{fig:other_datasets} top-row, two left-most examples), a type of knowledge for which current VLMs are powerful in both theory and practice~\cite{alayrac22neurips,chen22arxiv,hao22arxiv,yang22aaai}.
These datasets~\cite{marino19cvpr,schwenk22eccv} also include some questions requiring knowledge of detailed properties, but mostly about coarser basic-level categories~\cite{rosch:categorization78} (\eg Fig.~\ref{fig:other_datasets} bottom-left).
There are a few other datasets which target detailed properties~\cite{jain21sigir,shah19aaai}.
But both~\cite{jain21sigir,shah19aaai} lack attribution annotations, 
~\cite{jain21sigir} is very small (Tab.~\ref{tab:dataset_comparison} right-most column),
and~\cite{shah19aaai} focuses only on celebrities (Fig.~\ref{fig:other_datasets} fourth column).
Hence no current VQA dataset is fully satisfactory. %

\input{03_rw_comparison_table}        %

In this paper we introduce the \kivqa dataset, which offers several attractive features (Fig.~\ref{fig:our_dataset}, Tab.~\ref{tab:dataset_comparison}).
Our dataset asks questions about fine-grained categories from iNaturalist 2021~\cite{horn21cvpr} (Fig.~\ref{fig:our_dataset} bottom row) and instances from the Google Landmarks Dataset v2~\cite{weyand20cvpr} (Fig.~\ref{fig:our_dataset} top row).
We construct questions about detailed properties based on Wikipedia (\eg founder of building, maximum age of animal). 
As a consequence, all questions in our dataset are about \emph{encyclopedic} knowledge.
Furthermore, we provide a controlled knowledge base suited to answer these questions:
2M Wikipedia pages consisting of free-form text and images~\cite{srinivasan21sigir}.
We also mark ground-truth {\em attribution} for each answer at the granularity level of a section within a Wikipedia page.
Importantly, our dataset is collected at \emph{scale}: we have 221k unique question+answer pairs each matched with around 5 images, resulting in a total of 1M examples. 
This makes our dataset the largest of its kind. 
Finally, many of our questions are complex \emph{two-hop} questions~\cite{yang18emnlp}, which require multiple different documents from the knowledge base to solve (Fig.~\ref{fig:our_dataset} right).

We validate the usefulness of our dataset through several experiments.
In particular, we demonstrate that a large VLM (PaLI~\cite{chen22arxiv}) which yields state-of-the-art results on OK-VQA~\cite{marino19cvpr} performs poorly on our dataset ($13.0\%$ accuracy).
Next we demonstrate through an oracle experiment that retrieval-augmented models can yield $87.0\%$ accuracy.
Finally, we use an online image search engine to build an automatic retrieval-augmented prototype, reaching $48.8\%$ accuracy.
We conclude that
(1) our dataset poses a strong challenge in VQA and is in fact too hard for standard VLMs;
(2) retrieval-augmented VLMs demonstrate a strong potential for addressing encyclopedic knowledge;
(3) our results leave significant headroom for further research to improve retrieval-augmented VLMs.

%% file: 03_rw_comparison_table.tex
\begin{table*}[t]
    \centering
    \vspace{-.9cm}
    \begin{adjustbox}{max width=\textwidth}
    \begin{tabular}{lcccccc}
         \toprule
          & OK-VQA~\cite{marino19cvpr} & A-OKVQA~\cite{schwenk22eccv} & FVQA~\cite{wang18pami} & S3VQA~\cite{jain21sigir} & KVQA~\cite{shah19aaai} & \kivqa (ours) \\
         \midrule
         \emph{Truly multimodal}                         & ++ & ++ & ++ & ++ & ++ & ++ \\
         \emph{Encyclopedic}                             & $\pm$ & - & - & + & ++ & ++ \\
         $\quad\hookrightarrow$ Detailed properties           & + & $\pm$ & - & ++ & ++ & ++ \\
         $\quad\hookrightarrow$ fine-grained categories / instances & - & - & - & + & ++ & ++ \\
         \emph{Controlled knowledge base}                & - - & - - & + & $\pm$ & $\pm$ & ++ \\  
         $\quad\hookrightarrow$ answer supported by KB        & - - & - - & ++ & ++ & ++ & ++ \\
         $\quad\hookrightarrow$ KB provided                   & - - & - - & ++ & - - & ++ & ++ \\
         $\quad\hookrightarrow$ KB free-form                  & - - & - - & - - & - - & - - & ++ \\
         $\quad\hookrightarrow$ attribution                   & - - & $\pm$ & ++ & - - & - - & ++ \\
         \emph{Scale}                                    & $\pm$ & + & - - & - - & ++ & ++ \\
         \emph{Two-hop}                                & - - & - - & - - & - - & ++ & + \\
         \cmidrule(lr){1-7}
         Subject & various & various & various & various & celebrities & fine-grained species, landmarks \\
         Number of text questions $Q$               & 14k & 25k & 6k & 7k & 183k & 221k \\
         Number of images $I$                  & 14k & 24k & 2k & 7k & 25k & 514k \\
         Number of unique VQA triplets $(I,Q,A)$ & 14k & 25k & 6k & 7k & 183k & 1,036k \\
         \bottomrule
    \end{tabular}
    \end{adjustbox}
    \caption{\small \textbf{Comparison of recent VQA datasets.} We compare recent VQA datasets with our proposed dataset on VQA design principles.
    }
    \label{tab:dataset_comparison}
    \vspace{-.2cm}
\end{table*}

%% file: 02_design_principles.tex
\section{Design principles for our dataset}\label{sec:design_principles}

We create our VQA dataset with the following desired properties in mind:
(1) \emph{Truly Multimodal.}
The questions should not be answerable without the image or the textual question~\cite{goyal17cvpr}.
(2) \emph{Encyclopedic.}
The questions should be about detailed properties of fine-grained categories or instances; a type of questions which are problematic for vanilla VLMs~\cite{alayrac22neurips,chen22arxiv,hao22arxiv,yang22aaai}.
(3) \emph{Controlled knowledge base.}
Each answer in our dataset should be attributable to that specific part of the knowledge base which supports it. Hence the knowledge base is an integral part of our dataset, which enables measuring whether a model answers a question correctly for the right reason. For generality we want this knowledge base to be free-form text and to contain images.
(4) \emph{Scale.}
The dataset should be large. Dataset size has always mattered and this is even more true with the increasingly large VLM models (\eg~\cite{aghajanyan22arxiv,chen22arxiv,jia21icml,yuan21arxiv,radford21icml}).
(5) \emph{Two-hop.}
A portion of our questions should require knowledge 
from multiple different documents from the knowledge base. Including such complex two-hop questions~\cite{yang18emnlp} leaves substantial headroom for future model development.

%% file: 03_related_work.tex
\section{Related Work}\label{sec:related_work}

\para{Visual Question Answering (VQA).}
DAQUAR~\cite{malinowski14neurips}, FM-IQA~\cite{gao15neurips},
Visual Madlibs~\cite{yu15iccv},
VQAv1~\cite{antol15iccv} and VQAv2~\cite{goyal17cvpr} are early VQA datasets.
These datasets mostly ask visual questions that can be answered based on the query image and generic knowledge such as basic-level category recognition (\eg `cat'), counting items, colors, etc.

\para{Knowledge-based VQA.}
Tab.~\ref{tab:dataset_comparison} and Fig.~\ref{fig:other_datasets} compare datasets which are designed to require knowledge not present in the image~\cite{jain21sigir,marino19cvpr,schwenk22eccv,shah19aaai,wang18pami}.
In particular, OK-VQA~\cite{marino19cvpr} and A-OKVQA~\cite{schwenk22eccv} mostly require commonsense knowledge. In addition, some questions (18\%) in A-OKVQA do require knowledge of detailed properties, but about basic-level categories. Finally, 3\% of the questions require knowledge about physics.
In this paper we create a dataset with questions \emph{exclusively} about detailed properties of fine-grained categories and instances (Fig.~\ref{fig:our_dataset}).
We believe that such encyclopedic questions truly require access to a knowledge base to be answered.
In fact, we release the knowledge base along with our dataset, whereas no explicit
knowledge base was involved in the creation of~\cite{marino19cvpr,schwenk22eccv}.

Some existing VQA datasets are supported by a knowledge base~\cite{jain21sigir,shah19aaai,wang18pami}.
FVQA~\cite{wang18pami} is about commonsense knowledge.
KVQA~\cite{shah19aaai} asks detailed properties about celebrities.
Both~\cite{shah19aaai,wang18pami} are based on a structured knowledge base (RDF triplets).
Instead, our knowledge base is readily available free-form Wikipedia text, which has more information.
Furthermore, KVQA is exclusively about people, which requires specialized methods based on face detection and recognition. In contrast, our dataset offers a broader range of topics, including animals, plants and landmarks.
S3VQA~\cite{jain21sigir} is the most related work to ours. It builds on categories of OpenImages~\cite{kuznetsova20ijcv}, some fine-grained and others more basic-level.
They automatically generate questions from Wikipedia articles, which by construction are about detailed properties. Then they let experts manually select and rephrase relevant questions. 
Our paper also automatically generates QA pairs from Wikipedia.
However, we go beyond~\cite{jain21sigir} in several ways:
(1) our dataset is much larger (7k vs 1M VQA triplets),
(2) we include more complex multi-hop questions,
(3) we release the knowledge base along with the QA pairs. This also includes ground-truth answer attribution in the form of the supporting Wikipedia section.

The InfoSeek dataset~\cite{chen23arxiv} is concurrent to our work. It also targets detailed properties of fine-grained categories, but does not include multi-answer and two-hop questions, and was collected using a different annotation protocol.

\para{Automatic question generation.}
A few VQA datasets are generated from image captions~\cite{ren15neurips,changpinyo22naacl}. While this enables much larger scale, it results in simple visual questions. We build a portion of our dataset using a similar pipeline as~\cite{changpinyo22naacl}, but apply it to Wikipedia pages instead to obtain high-quality questions on detailed properties. As another difference, we verify them with human annotators.

\para{Multi-Hop Reasoning in NLP.}
The NLP community has many text-only QA datasets (\eg~\cite{dunn17arxiv,joshi17acl,khot20aaai,mihaylov18emnlp,rajpurkar16emnlp,talmor18naacl,welbl18acl,yang18emnlp}). Notably, HotpotQA~\cite{yang18emnlp} introduces the concept of `bridge entity' which links two related entities. They use this to show an annotator two related Wikipedia paragraphs from different pages and ask to create questions which requires knowledge from both paragraphs.
We use `bridge entities' to automatically chain two questions together into a compound two-hop question.

\para{Retrieval-augmented models.}
Our dataset seems particularly suited for retrieval-augmented models.
While pioneered in NLP~\cite{borgeaud22arxiv,guu20icml,khandelwal20iclr,lewis20neurips}, several recent works use retrieval for VQA.
KRISP~\cite{marino21cvpr} leverages triplets encoding facts, categorical knowledge and object relationships in a graph-based reasoning framework.
KAT~\cite{gui22naacl} uses Wikidata triplets, and a reasoning module cross-attends them with GPT-3 answers, the result being fed into a decoder for answer generation.
InFactuality~\cite{vickers21acl} leverages index images to link entities present in the query image and Wikidata triplets to gather facts, which are fed into a UNITER~\cite{chen20eccv} reasoning module.
REVEAL~\cite{hu23cvpr}'s external knowledge is composed of image-text pairs, question answering pairs and knowledge graph triplets, which are used to assist a generator module in producing answers.

%% file: 04_dataset.tex
\section{Dataset}
\label{sec:dataset}

We now detail the construction of our dataset while following the design principles (Sec.~\ref{sec:design_principles}). To achieve \emph{scale} we automate construction whenever possible. We use human annotators to ensure quality and to provide information which we could not get automatically. We simplify human annotation tasks as much as possible, which increases both the quality and efficiency of their work.

We define a unit for our VQA task as a triplet $(I,Q,A)$. The multi-modal question features an image $I$ and accompanying textual question $Q$. The subject of the question $Q$ appears in $I$ and is a core concept of our work. We refer to the category of the subject as $C$ (\eg `Horezu Monastery' or `Gila Monster' in Fig. \ref{fig:our_dataset}-left).
The answer $A$ is purely textual.
Our dataset also records the evidence for each answer in the knowledge base, as a section within the Wikipedia page for $C$ where the answer is found. 
In contrast to many existing VQA datasets, we record only a single answer, since it is unambiguous given the knowledge base.

To make our dataset \emph{truly multi-modal}, we always refer to the subject of the question by its super category (\eg `this monastery', `this reptile', as opposed to  `the Horezu Monastery', `the Gila monster', \etc).
This ensures visual recognition is required to solve our task, as there would be many potential answers to the textual question alone (\eg a different answer for each monastery in the world).

\begin{figure*}[th]
    \vspace{-1.1cm}
    \resizebox{\textwidth}{!}{
        \includegraphics{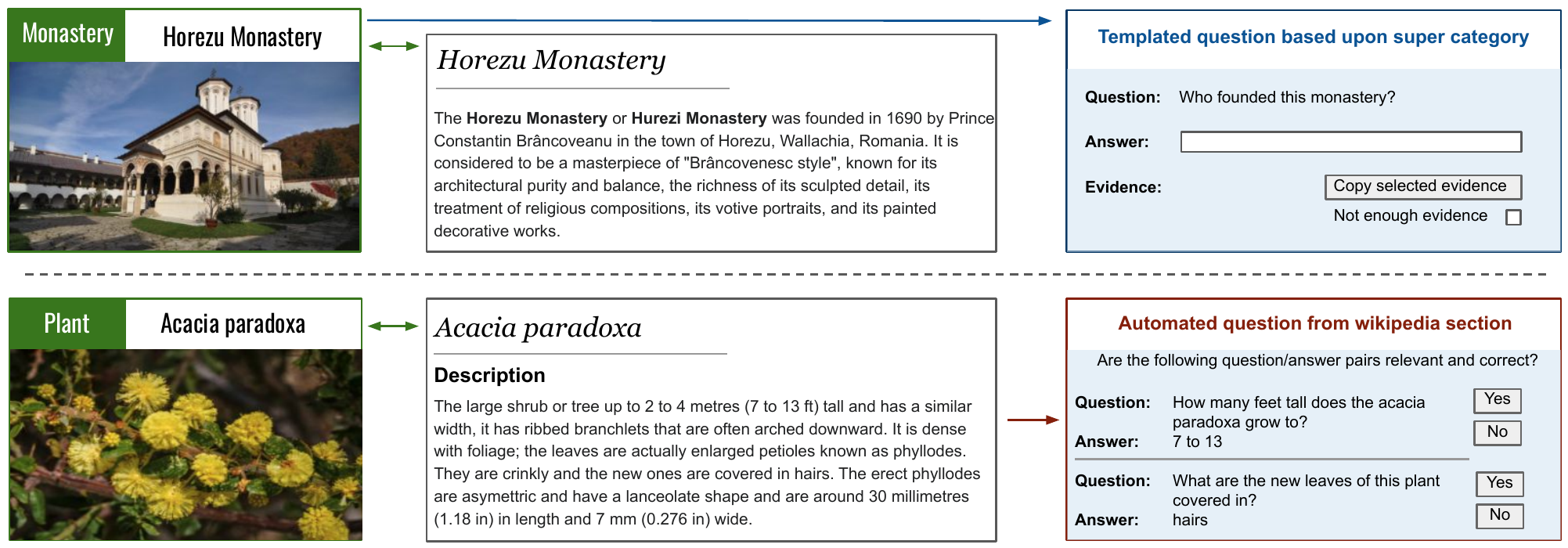}
    }
    \caption{\small \textbf{Data collection for templated and automatically generated single-hop questions.} 
    Top: Experts create templated questions $Q$ based on a super category (\eg Monastery). 
    Annotators are given the full Wikipedia page for a particular $C$, and asked to provide the answers $A$ and the evidence.
    Bottom: Questions are automatically generated from a Wikipedia section of $C$ and validated by human annotators.
    In both processes, the annotator never sees the image $I$ corresponding to $C$.
    }
    \label{fig:one_hop_pipeline}
    \vspace{-.0cm}
\end{figure*}

\subsection{Supporting Datasets}\label{sec:supporting_datasets}
We build on two of the largest existing datasets with fine-grained categories:
iNaturalist~2021 (iNat21)~\cite{horn21cvpr} and Google Landmarks Dataset V2 (GLDv2)~\cite{weyand20cvpr}.
\textbf{iNat21} is a fine-grained visual classification dataset containing 2.7M images depicting 10,000 species, grouped into 11 super categories: plants, insects, birds, mammals, \etc.
\textbf{GLDv2} is a landmark recognition dataset with 4M images depicting 200k landmarks. 
Each landmark is sourced from Wikimedia Commons~\cite{wikimediacommons}, enabling us to mine the existing category-hierarchy (allowing to define super categories, like bridges, castles, lighthouses, lakes \etc), and to link them to Wikipedia articles.
We use these provided annotations to speed up the annotation of our dataset, to identify relevant knowledge categories, and to assign images $I$ to textual questions $Q$ automatically.

We create our controlled knowledge base starting from the WIT dataset \cite{srinivasan21sigir}, which contains 37M image-snippet pairs with 11M unique images from Wikipedia.
We select all images linked to an English snippet and then extend WIT to include the full corresponding Wikipedia article (snapshot of 13 August 2022).
This spans 2M English articles.
We then identify which categories of iNat21 and GLDv2 map one-to-one to a single Wikipedia article. We found such unique mappings for 80\% of the iNat21 and 50\% of the GLDv2 categories. We only create questions for those categories with a uniquely identifiable Wikipedia article.

The combination of iNat21 \& GLDv2 with Wikipedia as knowledge base enables creating $(I,Q,A)$ triplets covering the long-tail: questions about detailed properties (on Wikipedia text) of fine-grained categories (species from iNat21) and instances (landmarks from GLDv2). Hence, this satisfies the \emph{encyclopedic} design principle.

\subsection{Single-hop questions}
\label{sec:single_hop}

The answer to a single-hop question can be found in the Wikipedia article of the category $C$ which is the subject of the question.
We construct such questions in two alternative ways: \emph{templated} and \emph{automatically generated}.

\para{Templated single-hop.}
We use of the super category annotations available in the supporting datasets (\eg reptiles for iNat21, monasteries for GLDv2).
For each super category we manually define several questions, like \emph{Who founded this monastery?}.
We then ask human annotators to answer these questions for a particular species/landmark $C$ (\eg Horezu Monastery), 
showing them the relevant Wikipedia page (see Fig. \ref{fig:one_hop_pipeline}).
Hence, the annotators do not have to recognize $C$ in an image, which would require expert knowledge.
Instead, they just have to identify whether the answer is in the Wikipedia article. 
If yes, they mark the evidence for it (for attribution). 
If no, we discard that question (as it cannot be answered with our knowledge base and it would break the \textit{controlled knowledge base} principle).
We start with 10 questions per category, resulting in $4.3$ questions on average per species with a valid answer for iNat21, and $3.4$ per landmark for GLDv2 (see examples in Fig.~\ref{fig:our_dataset} first column).

Now we have a textual question $Q$ and an answer $A$. To form a complete multi-modal question, we pair $Q$ with an image $I$ of $C$ from the supporting datasets.
Note that any $Q$ occurs multiple times in the dataset with different answers.
Hence by design the accompanying image $I$ is necessary to resolve the specific category $C$ and a model simply remembering \qapair pairs cannot solve the challenge.

\para{Automatically generated single-hop.}
We increase the diversity of questions in the dataset with automatically generated questions (Fig. \ref{fig:one_hop_pipeline}).
Similar to~\cite{jain21sigir}, we feed the Wikipedia article of a category to a question generation model~\cite{changpinyo22naacl}.
It processes a section of an article at a time, producing a large number of \qapair pairs (typically 100s),
by inverting statements from the text.

We include two filtering steps.
First, we require the category name of $C$ to be used in the question. 
This reduces the number of questions by $20\times$, and ensures that the question is about $C$ (and not about another entity found in the snippet of text).
Second, to increase diversity we remove near identical questions and limit the number of questions per Wikipedia section.

\begin{figure*}[t]
    \centering
    \vspace{-.0cm}
    \resizebox{\textwidth}{!}{\includegraphics{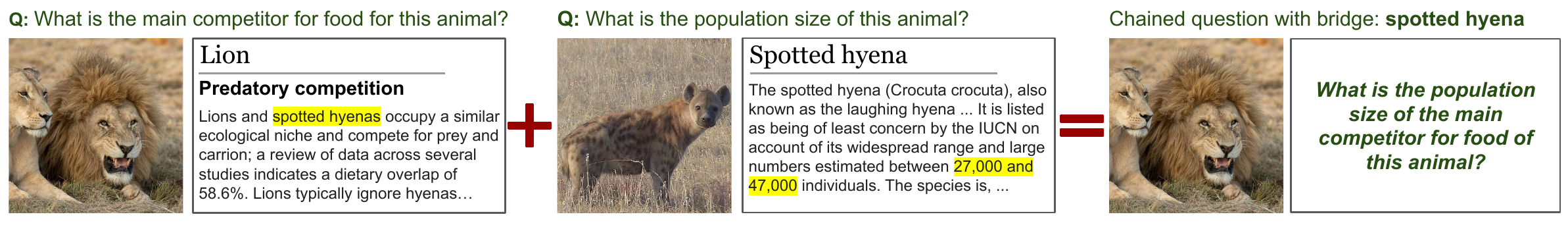}}
    \caption{\small %
        \textbf{Illustration of two-hop question generation using bridge entities.} 
        The bridge entity, \eg spotted hyena, is an answer to a single-hop question which has its own entry in the knowledge base.
        Hence we can ask a second single-hop question about it. 
        These two questions are then chained into a two-hop question.
    }
    \label{fig:two_hop_example}
\end{figure*}

The human annotation task is now extremely simple: validate whether the \qapair pair is relevant and correct, given the Wikipedia section used to generate it.
A question $Q$ is \textit{relevant} if it is is well-formed, makes sense semantically, and is expected to have an unambiguous answer. The answer $A$ is \text{correct} if it is supported by the given Wikipedia section.
Finally, we can directly use the Wikipedia section used to generate the QA pair as evidence for the answer (attribution for free!) 

About 80\% of the questions sent to annotators are validated positively. 
The questions contain (by design) the name of the category $C$, and we rephrase them using a carefully-prompted FLAN~\cite{chung2022scaling} version of PaLM~\cite{chowdhery22arxiv} to replace the mention of $C$ by its super category, which produces $Q$.
For example, the question \emph{Where is the Église Saint-Cannat located?} is rephrased by PaLM to \emph{Where is this church located?}
An image $I$ from $C$ from a supporting dataset is then associated to this question.

\para{Multi-answer questions.}
For many properties of a category $C$ the response should be a list with multiple answers. For example \emph{What fish can be found in this lake?} or \emph{Where is this bird found?}, could be answered with a list of fish (or countries, respectively; Fig.~\ref{fig:our_dataset}, third column). 
For these kinds of questions we define the multi-answer question type.
We filter the automatically generated QA pairs for questions likely to have multiple answers, and extract an initial list. Then we ask the annotators to complete and verify the list with all the possible answers.

\subsection{Two-hop questions}
\label{sec:multi_hop}
Two-hop questions are defined as requiring two reasoning steps to obtain the final answer.
Here we focus on \emph{chained} two-hop questions, which require two consecutive retrieval steps.
To construct such questions at scale, we use the notion of a \emph{bridge entity}~\cite{yang18emnlp}:
if the answer to a single-hop question is an entity with its own Wikipedia page, it serves as a bridge entity, about which we can ask a second single-hop question, see Fig.~\ref{fig:two_hop_example}.

We automatically identify bridge entities from the single-hop answers (Sec.~\ref{sec:single_hop}),
then manually discard words like \emph{yes, blue}, and \emph{heavy} which would yield very artificial multi-hop questions.
We then automatically generate and validate questions for these bridge entities using our automatically generated single-hop pipeline.

We create the final two-hop question automatically using PaLM, by feeding it the two single-hop questions with an appropriate prompt. Unfortunately, these two-hop questions are often incorrect.
Therefore, we create a second prompt to validate the two-hop question: PaLM is asked to answer the two-hop question using the two initial single-hop questions with answers as context. If the predicted answer is identical to the answer of the second single-hop question, then the two-hop question is validated and kept in our dataset; otherwise, it is discarded.
This increases the accuracy of the two-hop questions substantially.
Examples of validated two-hop questions are given in Fig.~\ref{fig:our_dataset}~(right).

\subsection{Dataset statistics}\label{sec:data_splits}
Our dataset contains in total 1M $(I,Q,A)$ triplets (Tab.~\ref{tab:dataset_statistics}).
These are derived from a total of 221k textual \qapair pairs from 16.7k different categories $C$.
Each \qapair pair is combined with (up to) 5 images $I$ showing different instances of the same category $C$ from the supporting datasets.
This increases visual diversity and in total we use 514k unique images.
There are in total 15k textual single-hop templated questions, and 158k automatically generated questions.
Moreover, the dataset contains 25k multi-answer questions and 22k two-hop questions.

Looking in detail at the 221k \qapair pairs, 175k of them have a unique $Q$ (\ie no string match with any other question).
Moreover, there are 95k unique answers, of which 73k occur only once. The remaining 22k answers follow a Zipf's law: 11k answers occur twice, 2k answers occur $10+$ times, and less than 400 answers occur more than 50 times.

\input{04_dataset_table_small}%

\para{Train / Val / Test splits.}
To allow for evaluating different properties of the dataset, we carefully design its train/val/test splits.
First, there is no overlap in the images $I$ used in the train, val, or test splits.
For GLDv2, we sample images for all our splits from their train-clean split, which avoids most noisy labels in their dataset.
For iNat21, we sample our train images from their train set, and our val and test images from their validation set (their test set annotations are not publicly available).
Next, about two-thirds of the questions $Q$ and half of the answers $A$ in our val and test splits do not occur in our train split.
Finally, roughly 17\% of the subject categories $C$ in our val and test splits are not used in our train split.
This allows for analyzing different desirable properties of VQA models on our dataset.

\input{A0_diversity}

%% file: 04_dataset_table_small.tex
\begin{table}[b]\centering
\vspace{0.1cm}
\resizebox{\columnwidth}{!}{
\begin{tabular}{lrrrrrrr}\toprule
&\multicolumn{3}{c}{\textbf{Number of (Q, A) pairs}} & 
&\multicolumn{3}{c}{\textbf{Total (I, Q, A) triplets}}\\\cmidrule{2-4}\cmidrule{6-8}
&\textbf{Train} &\textbf{Val} &\textbf{Test} & &\textbf{Train} &\textbf{Val} &\textbf{Test}\\\midrule
Templated      &         13,928 &            400 &          1,000 &&         66,535 &          1,827 &          1,000 \\
Automatic      &        153,441 &          1,750 &          2,750 &&        737,114 &          8,025 &          2,750 \\
Multi Answer   &         23,929 &            400 &          1,000 &&        112,736 &          1,844 &          1,000 \\
Two Hop        &         21,040 &            400 &          1,000 &&         99,866 &          1,895 &          1,000 \\[1mm]\midrule
\textbf{Total} &\textbf{212,338}&\textbf{  2,950}&\textbf{  5,750}&&\textbf{1,016,251}&\textbf{ 13,591}&\textbf{ 5,750}\\
\bottomrule
\end{tabular}
}
\caption{\small \textbf{Dataset Statistics.} We report the number of (question, answer) pairs, and (image, question, answer) triplets for different question types. In total our dataset contains 1M VQA triplets, making it the largest of its kind.}
\label{tab:dataset_statistics}
\end{table}

%% file: A0_diversity.tex
\begin{table}[t]
    \centering
    \vspace{-.3cm}
    \begin{adjustbox}{max width=\columnwidth}
    \begin{tabular}{lHHrr}
    \toprule
       {\bf Dataset} & {\bf Knowledge} & {\bf dataset size} & {\bf variety of $Q$} & {\bf disparity of $Q$} \\
       & {\bf based} & total number of train questions & \# unique bigrams & avg. cosine dist. \\
       \midrule
       FVQA~\cite{wang18pami} & \checkmark & 6k & 7.9k & 0.620 \\
       KVQA~\cite{shah19aaai} & \checkmark & 183k & 12.8k & 0.504 \\
       OK-VQA~\cite{marino19cvpr} & \checkmark &  9k & 19.3k & 0.843 \\
       S3VQA~\cite{jain21sigir} & \checkmark & 5k & 19.7k & 0.805 \\
       A-OKVQA~\cite{schwenk22eccv} & \checkmark & 17k & 33.9k & 0.856 \\
       \evqa & \checkmark & 212k & 257.9k & 0.833 \\
       \bottomrule
    \end{tabular}
    \end{adjustbox}
    \caption{\small \textbf{Diversity of questions $Q$} measured in terms of \emph{variety} and \emph{disparity} (higher is better).
    Disparity numbers are mostly taken from~\cite{schwenk22eccv}, but we reproduced their A-OKVQA result to validate our re-implementation.
    Our dataset has the largest variety and is close to the best ones on disparity.}
    \label{tab:disparity}
\end{table}

\subsection{Question Diversity}
\label{sec:diversity}

We want to compare the diversity of our VQA questions to other datasets.
However, diversity is a broad concept which manifests through a combination of three basic properties:
variety, disparity, and balance~\cite{stirling07interface}.
Here we focus on variety and disparity.

\para{Variety.} 
This refers to the semantic variety spanned by the VQA questions.
One simple measure of variety is the number of unique textual questions $Q$ (175k in our dataset). However, this ignores the fact that some questions are semantically similar.
Another measure is the number of topics covered by the questions, for which we can use the number of categories $C$ (16.7k in our dataset).
However, each dataset typically uses a different set of categories at different levels of granularity, making their category counts hard to compare directly.
So instead, we consider the number of unique bigrams across all questions $Q$, as a reasonable approximation of the semantic space spanned by all $Q$.

\para{Disparity.}
This measures how different the questions are from each other.
We follow the measure introduced in A-OKVQA~\cite{schwenk22eccv}:
(1) we first project all questions $Q$ into a common semantic space using a publicly available~\cite{multi-qa-minilm} Sentence-BERT model~\cite{reimers19emnlp}; then
(2) we measure the average cosine distance between all question pairs (defined as $1 - $ cosine similarity by~\cite{schwenk22eccv}).
Note that while~\cite{schwenk22eccv} proposed this as a generic measure of diversity, it actually only measures \emph{disparity} as defined in~\cite{stirling07interface}.

\para{Results.}
We report variety and disparity for knowledge-based VQA datasets in Tab.~\ref{tab:disparity}.
Our dataset offers the largest question variety by an order of magnitude. Importantly, this is not only due to dataset size: while \evqa and KVQA~\cite{shah19aaai} are the largest datasets and have a comparable number of questions, KVQA has much less variety.
In terms of disparity, our dataset is close to the best ones OK-VQA~\cite{marino19cvpr} and A-OKVQA~\cite{schwenk22eccv}, and better than FVQA, KVQA and S3VQA.

%% file: 05_experiments.tex
\section{Experiments}
\label{sec:experiments}

We evaluate PaLI~\cite{chen22arxiv}, PaLM~\cite{chowdhery22arxiv} and GPT-3~\cite{brown2020language} on our dataset.
We apply all models directly and in retrieval-augmented settings.
For evaluation we use the test split of our dataset on the single-hop templated and automatically generated questions (except for Sec.~\ref{sec:experiments_difficulty} where we also evaluate multi-answer and two-hop questions).
We measure accuracy as the percentage of questions where the predicted model answer matches the ground-truth answer.
All ground-truth answers and model predictions are pre-processed following standard VQA practices~\cite{goyal17cvpr} to facilitate checking their correctness (remove articles, punctuation, etc.).
Furthermore, we use the 'BERT Matching' criterion BEM~\cite{bulian2022tomayto} to determine whether a predicted answer is correct given the question and the ground-truth answer. BEM evaluation allows for more flexibility in the answer formulation than classical exact matching,
coming much closer to human judgement of correctness,
as is shown in~\cite{bulian2022tomayto} and in our user study in Appendix~\ref{sec:bem_quality}.
We consider an answer correct when its BEM score is $\geq 0.5$.
If an example has multiple ground-truth answers, we compute the BEM score for each and choose the maximum.

\subsection{Large Models without Retrieval}
\label{sec:large_vlms}

\para{PaLI.}
To measure the performance of a large VLM on \kivqa we use PaLI~\cite{chen22arxiv}.
It yields the state-of-the-art accuracy on OK-VQA~\cite{marino19cvpr} ($64.5\%$), and outperforms retrieval-augmented models such as KAT~\cite{gui22naacl} ($53.1\%$) and REVEAL~\cite{hu23cvpr} ($59.1\%$) by a large margin (Tab.~\ref{tab:okvqa_results}).
This demonstrates that the type of knowledge required to solve OK-VQA can be captured by large VLMs.
PaLI is therefore a good candidate to verify whether large VLMs capture also encyclopedic knowledge. 
We use a PaLI model with 17B parameters pre-trained on a huge amount of data including Wikipedia (so it has seen the knowledge base containing the answers for our dataset).
Our PaLI model is additionally fine-tuned on OK-VQA and therefore particularly suited for VQA tasks.
We feed this model $Q,I$ inputs for each test sample to produce model answers.

\begin{table}[t]
    \vspace{-.3cm}
    \centering
    \resizebox{0.8\linewidth}{!}{
    \begin{tabular}{c|c|c}
        \toprule
        \textbf{Model} & \textbf{Retrieval} & \textbf{OK-VQA Accuracy} \\
        \midrule
        KAT~\cite{gui22naacl} & \checkmark & 53.1 \\ 
        REVEAL~\cite{hu23cvpr} &  \checkmark &  59.1 \\ 
        PromptCap~\cite{hu2022promptcap} & - & 60.4 \\
        PaLI~\cite{chen22arxiv} & - & \textbf{64.5} \\ 
        \bottomrule
    \end{tabular}
    } %
    \caption{\small \textbf{OK-VQA:} PaLI is state-of-the-art on OK-VQA. It is therefore an good candidate to apply to our \kivqa.}
    \label{tab:okvqa_results}
\end{table}

\input{05_table_single_hop}

Tab.~\ref{tab:kivqa_results} (first row) shows that PaLI has an accuracy of only $13.0\%$.  
This low performance indicates that PaLI either fails to recognise fine-grained categories and instances, or fails to provide detailed properties, or both. Hence encyclopedic knowledge is truly difficult for this model.

\para{PaLM and GPT-3.}
Models with a larger language understanding component~\cite{brown20arxiv,brown2020language,chowdhery22arxiv,rae22arxiv} might be better at memorizing detailed properties and extracting information from textual knowledge base entries.
Therefore, we experiment with PaLM 2~\cite{chowdhery22arxiv} (\texttt{text-bison@001} model, available through the PaLM API) 
and GPT-3~\cite{brown2020language} (\texttt{text-davinci-003}, accessible through the OpenAI API).
Both are trained on massive amounts of text including Wikipedia. GPT-3 is especially large, with 175B parameters.
Here we feed each model only the text questions $Q$ as thery cannot consume images. This measures how well our dataset can be solved by language alone.
These models reach a modest accuracy  ($19.7\%, 15.5\%$, Tab.~\ref{tab:kivqa_results}).
While they only takes textual inputs, they can be made multi-modal by adding a visual retrieval mechanism, as we do in the following sections.

\subsection{Large Models with Oracle Retrieval}
\label{sec:oracle_xp}

\para{Oracle retrieval of subject $C$.} 
This experiment tests whether the most difficult aspect of \kivqa is recognizing
fine-grained categories and instances in the image. 
Therefore we provide along with the test question its corresponding subject category $C$ as an additional textual input to our models (\ie as part of the prompt; \emph{Oracle - Subject $C$} in Tab.~\ref{tab:kivqa_results}).
While PaLI shows small improvements over its non-retrieval version, PaLM and GPT-3 improve considerably to $31.0\%$ and $26.9\%$, respectively.
However, this is still rather low performance.
We conclude that determining the fine-grained category or instance is only part of what makes our dataset difficult.

\para{Oracle retrieval of KB article.}
We now go a step further and provide the full ground-truth Wikipedia article about $C$ to the models in the prompt (\emph{Oracle, KB Article} in Tab.~\ref{tab:kivqa_results}).
This time, results increase dramatically: $29.7\%$ accuracy for PaLI, $78.4\%$ for PaLM and $77.4\%$ for GPT-3.
This demonstrates that
(1) memorizing detailed properties is a hard challenge for vanilla large VLMs,
(2) adding a retrieval component shows great potential for predicting detailed properties, and
(3) since PaLM and GPT-3 work much better than PaLI, having a strong language understanding model is important for extracting the right information from the free-form text KB article.

\para{Oracle retrieval of KB section.}
Finally, to understand the importance of retrieving information that is more precisely localized than an entire Wikipedia page, we provide to the models the ground-truth section which supports the answer (\emph{Oracle - KB Section} in Tab.~\ref{tab:kivqa_results}).
Again, accuracy improves to $48.8\%$ for PaLI, $87.0\%$ for PaLM and $82.1\%$ for GPT-3.
This demonstrates the importance of providing exact information to the language component. 
As a bonus, the smaller the retrieved document is upon which the model bases its answer, the more verifiable and interpretable this answer is.

\subsection{Large Models with Visual Retrieval}
\label{sec:retrieval_xp}

\para{Lens-based retrieval of KB article.}
To go beyond the oracle demonstration above, as proof-of-concept we propose to augment large models with a real retrieval system based on Google Lens~\cite{googlelens}.
Google Lens is an image retrieval system which indexes a huge amount of web images.
Given a query image, it finds other images based on their visual similarity and relevance to objects it recognizes in the query image.
It returns the most similar indexed images along with an entity prediction based on these top-ranked images.
To augment our system, we send Google Lens the query image $I$ to obtain its entity prediction.
We then find the best matching KB article for this entity in our knowledge base.
Finally, we feed the retrieved KB article as prompt to our models as in Sec.~\ref{sec:oracle_xp}.

Results are shown as \emph{Lens - KB Article} in Tab.~\ref{tab:kivqa_results}.
All models greatly outperform their non-retrieval augmented versions. PaLM and GPT-3 in particular more than double, to 48.0\% and 44.9\% respectively.

\para{Lens-based retrieval of KB section.}
Given a retrieved KB article, we now aim to select the most relevant KB section within it based on $Q$.
To do so we query PaLM with a special prompt feeding one KB section at a time, along with $Q$ while asking `can the answer to this question be found in this text?'.
We retain all sections for which PaLM answers `Yes' (usually only one), and consider their concatenation as the final retrieved `KB section'.
Finally, we feed this retrieved section in the prompt to all models (PaLI, PaLM or GPT-3) as in Sec.~\ref{sec:oracle_xp}.

Results are shown as \emph{Lens - KB Section} in Tab.~\ref{tab:kivqa_results}.
Performance is now even higher for PaLI ($28.1\%$) and PaLM ($48.8\%$), whereas GPT-3 does not seem to benefit further.
These numbers are roughly halfway between versions without retrieval and with oracle retrieval.
We conclude that retrieval augmentation works in practice, and that our dataset leaves significant headroom for future research on retrieval-augmented VLMs.

\para{CLIP-based retrieval.}
KAT~\cite{gui22naacl} and REVIVE~\cite{lin22neurips} are two retrieval-augmented VQA systems which use frozen CLIP~\cite{radford21icml} embeddings to perform retrieval.
In Appendix \ref{sec:suppmat_clip_comparison} we explore there whether KAT and REVIVE could potentially succeed on \evqa.

\subsection{Attribution for Retrieval-Augmented models}
\label{sec:attribution}

The attribution annotations of our dataset enable measuring whether retrieval-augmented models give the correct answer for the right reason.
Tab.~\ref{tab:attribution_results} measures attribution for the PaLM models with Lens retrieval.

First, we observe that Google Lens retrieves the correct KB Article $47.4\%$ of the time.
Furthermore, we report $45.6\%$ retrieval accuracy at the finer level of a KB Section.
This demonstrates that given a correctly retrieved KB Article, our system almost always finds the correct Section within it.
More importantly, if the retrieved KB Section is incorrect, accuracy drops drastically to $20.7\%$, close to the non-augmented variant (both when using PaLM).
In contrast, if the correct KB Section is found, accuracy is $82.3\%$. 
Thus, retrieving the correct document, and even better the correct section, is essential to performing well on our dataset.

\input{05_table_attribution}

\subsection{Comparison to PromptCap}
\label{sec:promptcap}

We compare our retrieval-based methods to PromptCap~\cite{hu2022promptcap}, which has strong performance on OK-VQA (Tab.~\ref{tab:okvqa_results}).
PromptCap is a system with multiple components.
It has a captioning model which consumes both $I$ and $Q$ and generates an image caption tailored to answer the question $Q$. This caption is then passed to GPT-3 as context to answer $Q$. Additionally, PromptCap performs in-context learning: for a given test question, it uses the CLIP embeddings~\cite{radford21icml} of $Q$ and $I$ to find the 32 nearest training examples. Then it includes their 32 corresponding PromptCap captions, questions and answers in the GPT-3 prompt.
In~\cite{hu2022promptcap} the underlying LLM was GPT-3. In this experiment we also apply PromptCap with PaLM and PaLI.
Results in Tab.~\ref{tab:kivqa_results} show that PromptCap indeed does better than using PaLI, PaLM, or GPT-3 alone. However, it still performs substantially worse than our retrieval-augmented methods. This further confirms the hard challenge our dataset poses.

\subsection{Multi-answer and two-hop questions}
\label{sec:experiments_difficulty}

To evaluate multi-answer questions, we convert the model prediction into a set of strings and compute the intersection-over-the-union (IoU) between this set and the set of ground-truth answers.
If IoU $>=0.5$ then we consider that prediction as correct. If not, then we use BEM to determine the equivalence of the prediction list string and the ground-truth list.
For two-hop questions, we use BEM as described for single-hop templated and automatically generated questions.

PaLI with Lens KB Section Retrieval obtains an accuracy of $9.2\%$ for multi-answer questions and $14.7\%$ for two-hop questions. 
Similarly, PaLM with KB Section retrieval achieves $33.6\%$ and $22.8\%$ for multi-answer and two-hop questions, respectively.
GPT-3 obtains $32.1\%$ for multi-answer and $18.7\%$ for two-hop questions, in between PaLI and PaLM.
These modest performances, especially on two-hop questions, confirm the challenge of our proposed tasks, and highlight the usefulness of \kivqa to measure progress on
designing retrieval mechanisms for such complex types of questions.

%% file: 05_table_single_hop.tex
\begin{table*}[t]
    \vspace{-.0cm}
    \centering
    \begin{adjustbox}{max width=1.4\columnwidth}
    \begin{tabular}{ccccrr}
    \toprule
    & & & \multicolumn{3}{c}{\textbf{Model}}  \\ 
    \cmidrule(lr){4-6}
    System & Retrieval & Extra input & \textbf{PaLI~\cite{chen22arxiv}}   & \textbf{PaLM~\cite{chowdhery22arxiv}} & \textbf{GPT-3~\cite{brown2020language}} \\
    \cmidrule(lr){1-6}
    Vanilla model                       & - &      -              & 13.0\% & 19.7\% & 15.5\%   \\
    \cmidrule(lr){1-6}
    \multirow{3}{*}{Oracle} & Subject $C$ & -         & 16.7\% & 31.0\% & 26.9\%    \\
                            & KB Article  & -         & 29.7\% & 78.4\% & 77.4\%    \\
                            & KB Section  & -        & 48.8\% & 87.0\% & 82.1\%    \\
    \cmidrule(lr){1-6}
    \multirow{2}{*}{Lens}   & KB Article  & - & 21.4\% & 48.0\% & 44.9\%    \\
           & KB Section & - & 28.1\% & 48.8\% & 44.6\%    \\ 
    \bottomrule
    \toprule
    PromptCap   & - & \makecell{Captions \\ NN train samples}     & 17.8\% & 29.7\% & 25.6\%   \\
    \bottomrule
    \end{tabular}
    \end{adjustbox}
    \caption{\small \textbf{Accuracy on single-hop questions.} We report model accuracy for our single-hop templated and automatically generated questions. While large models struggle with our \kivqa dataset, our experiments show the promise of augmenting them with a retrieval mechanism.
    }
    \label{tab:kivqa_results}
    \vspace{-.0cm}
\end{table*}

%% file: 05_table_attribution.tex
\begin{table}[t]
    \vspace{-.0cm}
    \centering
    \begin{adjustbox}{max width=\columnwidth}
    \begin{tabular}{ccccc}
    \toprule
    & \multicolumn{3}{c}{\textbf{Accuracy}} & \\ 
    \cmidrule(lr){2-4}    
    & & w/ correct & w/ incorrect & \textbf{KB retrieval}\\
    \textbf{Lens Retrieval} & Overall & retrieval & retrieval & \textbf{accuracy} \\ \midrule    
    KB Article & 48.0\%            & 77.7\%            & 21.2\%               & 47.4\%        \\
    KB Section & 48.8\%            & 82.3\%            & 20.7\%               & 45.6\%        \\ 
    \bottomrule
    \end{tabular}
    \end{adjustbox}
    \caption{\small \textbf{Attribution for retrieval-augmented VLMs}. We report model accuracy for PaLM~\cite{chowdhery22arxiv} conditional on retrieval results. Retrieval success rates are similar for articles and sections, but PaLM has better accuracy when augmented with sections.}
    \label{tab:attribution_results}
    \vspace{-.2cm}
\end{table}

%% file: 06_conclusions.tex
\section{Conclusions}
\label{sec:conclusions}

We introduced \kivqa, a large-scale VQA dataset about detailed properties of fine-grained categories and instances, which includes a knowledge base with answer attributions.
We demonstrated that our dataset is truly difficult for standard VLMs.
Additionally, we showed with both an oracle experiment and a prototype automatic system that augmenting these models with a retrieval component substantially improves results, yet leaving headroom for even further improvements.
Therefore our dataset enables future research on retrieval-augmented VLMs.

%% file: A1_user_study.tex
\section{Correctness of our dataset}
\label{sec:user_study}

We rely on iNat21 and GLDv2 to have clearly identifiable categories on the test images of our dataset, and to obtain their ground-truth labels.
Furthermore, we only use categories which unambiguously map to Wikipedia articles by explicitly seeking for one-to-one mappings (Sec.~\ref{sec:supporting_datasets}).
The main remaining possible source of error is whether a question can be answered given the corresponding Wikipedia article, and whether the recorded ground-truth answer is indeed correct. Human annotators systematically validate these aspects for every single question+answer pair (Sec.~\ref{sec:single_hop} and~\ref{sec:multi_hop}).
Yet, we perform here an additional user study to get an numerical estimate of quality in this sense.

In this study, we randomly sample 100 questions from the test set.
Then we ask six experts to each answer 50 questions each, given the corresponding Wikipedia page as reference. These expert were not involved in the original data collection process.
This process results in three answers per question.
If the majority of the 3 expert answers matches our collected ground-truth according to BEM~\cite{bulian2022tomayto}, we consider that question to be \emph{answerable} given the Wikipedia page, and our ground-truth to be correct.
We find that this holds for most of our questions (86\%).

To put this number in context, we also estimate the answerability of A-OKVQA~\cite{schwenk22eccv} with respect to their evaluation metric.
A-OKVQA follows previous VQA datasets and provides 10 ground-truth answers per question.
A predicted answer is counted as correct if it matches with 3 ground-truth answers (exact string matching~\cite{antol15iccv,schwenk22eccv}, after normalization such as removing punctuation and articles, converting numbers to digits, etc.).
So we can consider a question is answerable if there are at least 3 equivalent answers out of the 10.
We find that 86\% of the A-OKVQA questions are answerable.

To conclude, this user study demonstrates that our Encyclopedic-VQA dataset is of very high quality.

\section{Quality of BEM~\cite{bulian2022tomayto} evaluation measure}
\label{sec:bem_quality}

We want to understand how well the BERT Matching (BEM)~\cite{bulian2022tomayto} evaluation measure mirrors human judgments. To do so we build on the user study above and ask an expert human to judge whether each answer from the user study matches the collected ground-truth answer or not. The judge has access to both the question $Q$ and the corresponding Wikipedia page.
We find that the BEM judgements equal human judgements in the vast majority of the cases (96\%).
BEM is a stricter judge in 3\% of the cases, the human in 1\%.
This demonstrates that BEM correlates very highly with human judgement of correctness of an answer, confirming what reported in~\cite{bulian2022tomayto} for other datasets.

\para{Exact Match.}
Many VQA datasets perform exact string matching for their evaluation~\cite{antol15iccv,goyal17cvpr,marino19cvpr,schwenk22eccv}.
Because it is a strict evaluation measure and because their questions are more open-ended (they are not supported by a controlled knowledge base, Sec.~\ref{sec:dataset}), those datasets collect 10 ground-truth answers per question, to cover some of the variability in answer formulation. This then enables a more relaxed evaluation, as the model answer need only match some of
the 10 ground-truth ones (three matches for a perfect answer).
Nevertheless, it is instructive to verify whether exact matching would work on our dataset, where we have  a single ground-truth answer. We start from the publicly available exact matching implementation of~\cite{gui22naacl} and include additional relaxations for number comparisons (e.g. numbers over 100 - usually years - may be off by one, ranges are correct if they partially overlap, etc). We find that exact match judgements are equivalent to human judgements only in 68\% of the cases. Where they differ, exact match is always overly strict, rejecting answers that a human would judge as correct.
This demonstrates that exact matching does not work well for our dataset, and justifies our choice of BEM as the evaluation measure to check whether predicted model answers match ground-truth ones in Sec. \ref{sec:experiments}.

%% file: A2_clip_retrieval.tex
\section{CLIP retrieval in existing retrieval-augmented VQA systems~\cite{gui22naacl,lin22neurips}}
\label{sec:suppmat_clip_comparison}

\begin{table}[b]
    \centering
    \begin{adjustbox}{max width=\columnwidth}
    \begin{tabular}{crrrr}
    \toprule
    & \multicolumn{4}{c}{\textbf{Recall}}  \\ 
    \cmidrule(lr){2-5}
    Method & @1 & @5 & @10 & @20 \\ 
    \cmidrule(lr){1-5}
    CLIP~\cite{radford21icml}   & 3.3\%  & 7.7\%  & 12.1\% & 16.5\%   \\
    Lens~\cite{googlelens}      & 47.4\% & 62.5\% & 64.7\% & 65.2\%  \\
    \bottomrule
    \end{tabular}
    \end{adjustbox}
    \caption{\small \textbf{Recall results for CLIP and Lens.} We report the recall in retrieving the right KB article within the top-K documents.}
    \label{tab:suppmat_clip_recall}
\end{table}

KAT~\cite{gui22naacl} and REVIVE~\cite{lin22neurips} are two retrieval-augmented VQA systems which use frozen CLIP~\cite{radford21icml} embeddings to perform retrieval. More specifically, they first encode their text-only knowledge base (extracted from Wikidata) using the language tower of CLIP. At test time, given a VQA question, they encode its image $I$ with CLIP and then compare its embedding to the knowledge base embeddings to retrieve the most similar entries. The top few most similar entries are then passed on to a T5 model~\cite{raffel21jmlr} which is trained to produce the final answer.
Note how correct retrieval is crucial to perform well on \evqa, as the retrived knowledge base entries should contain the answer for the overall system to succeed. Hence, we now test the CLIP-based retrieval component of KAT~\cite{gui22naacl} and REVIVE~\cite{lin22neurips} in isolation, to estimate whether they would be able to succeed on \evqa.

We represent Wikipedia articles as the CLIP embedding of their title and description strings concatenated. At test time, given a VQA question, we use CLIP to embed the image $I$ to form the query for retrieval, proceeding as in~\cite{gui22naacl,lin22neurips}.
Results in Tab.~\ref{tab:suppmat_clip_recall} demonstrate that recall is low:
the correct Wikipedia article corresponding to the subject of the question is retrieved in the first position only 3.3\% of the time. Even within the top-20 results, retrieval accuracy is still only at 16.5\%. This suggest that KAT/REVIVE would not work well on our dataset.

\para{Lens retrieval.}
In Sec. \ref{sec:retrieval_xp} we use Google Lens to identify the subject $C$ of a VQA question in our datatset, i.e. a iNat21 or GLDv2 category, which works well (Tab.~\ref{tab:suppmat_clip_recall}).
Lens is an image-based system which indexes billions of images on the web and retrieves relevant ones based on their visual similarity. Hence, it has a greater chance to find a web image resembling a VQA test image, than when restricting the search only to Wikipedia. Attached to the image is often various meta-data which enables to determine the name of the subject category (which we then use to find the right Wikipedia page).
Recognizing landmarks and species from the natural world are typical use-cases so we can expect Lens to work well for recognizing the subjects $C$ in our dataset.
However, academic results on iNat21~\cite{inaturalistChallenge2021} and GLDv2~\cite{googleUniversalEmbeddingChallenge} suggests that specialized systems work very well on these datasets, suggesting that a good image-based classifier would make a viable substitute for Lens.

Generally, the core points of our paper are that (1) \evqa poses a hard challenge for large LLMs and VLMs, and (2) solving it requires augmenting the LLM/VLM with a retrieval component to access a knowledge base. The exact choice of retrieval component is flexible and likely subject to further exploration.

%% file: A3_qualitative.tex
\section{Qualitative results}

\begin{figure*}[t]
    \centering
    \includegraphics[width=\linewidth]{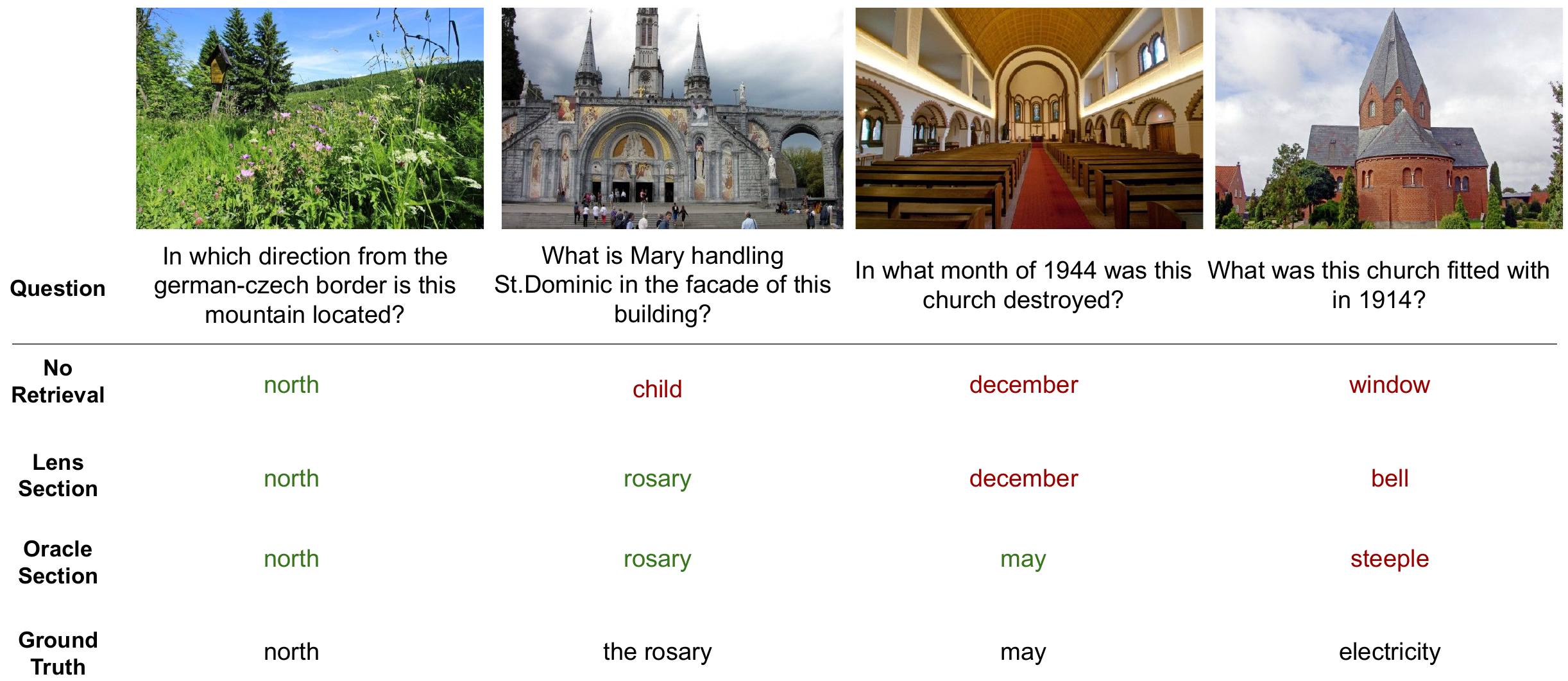}
    \caption{\small
    \textbf{PaLI qualitative results.}
    We present $4$ multi-modal questions and the answers produced with different experimental setups, using the \textbf{PaLI} model: without retrieval (``No Retrieval''), Lens-based retrieval of KB section (``Lens Section''), Oracle retrieval of KB section (``Oracle Section'').
    Additionally, we provide the ground-truth answer in the last row.
    We show cases where each of the setups can produce the correct answer, as well as an example where all methods fail (the right-most one).
    }
    \label{fig:supp_pali_qualitative}
\end{figure*}

\begin{figure*}[t]
    \centering
    \includegraphics[width=\linewidth]{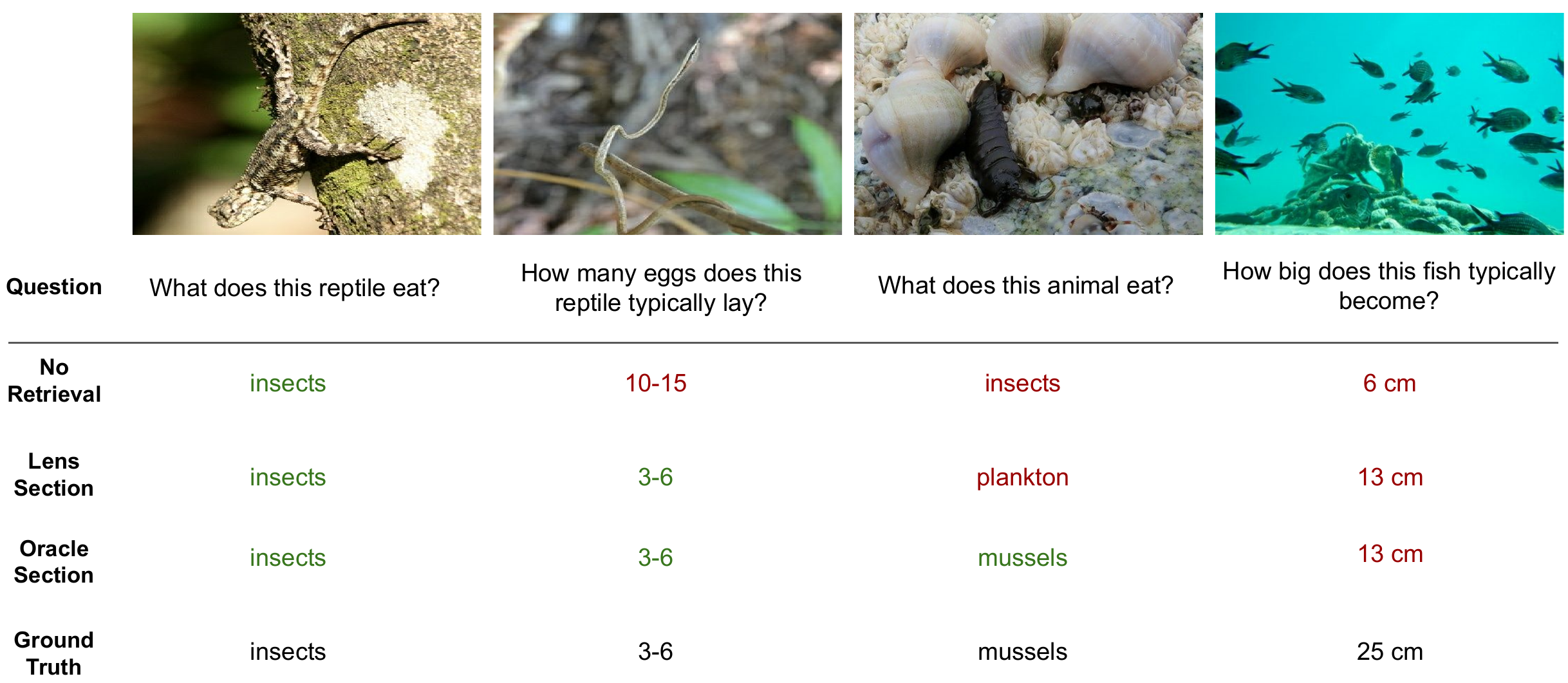}
    \caption{\small
    \textbf{PaLM qualitative results.}
    We present $4$ multi-modal questions and the answers produced with different experimental setups, using the \textbf{PaLM} model: without retrieval (``No Retrieval''), Lens-based retrieval of KB section (``Lens Section''), Oracle retrieval of KB section (``Oracle Section'').
    Additionally, we provide the ground-truth answer in the last row.
    We show cases where each of the setups can produce the correct answer, as well as an example where all methods fail (the right-most one).
    }
    \label{fig:supp_palm_qualitative}
\end{figure*}

We present qualitative results for some of the methods we experimented with in Figures \ref{fig:supp_pali_qualitative} and \ref{fig:supp_palm_qualitative}.
These illustrate cases where the different setups may work, and cases where all of them fail.
For example, given a textual question \emph{What does this reptile eat?}, PaLM can reasonably guess an answer without any additional context: \emph{insects} -- since that's what many reptiles eat.

For more specific information, relating to more detailed properties (\eg, number of eggs a specific reptile lays, dates, how big a specific fish may become), large models may still make reasonable guesses, but generally this leads to incorrect answers.
Augmenting large models with context from retrieved knowledge improves the accuracy in these cases substantially, leading to precise answers that are attributable to the piece of knowledge that was retrieved.

Finally, we also see cases where the generated answer is incorrect, even if the correct piece of knowledge is retrieved.
This indicates that in some cases large models may still have difficulties using retrieved knowledge to generate accurate answers.

%% file: A4_dataset_statistics.tex
\section{Dataset Statistics}

\input{04_dataset_table}      %

In \autoref{tab:dataset_statistics_appendix} we provide a more detailed overview of the dataset statistics. For more details see Section 4 of the main paper.

%% file: 04_dataset_table.tex
\newcommand{\datasetname}[1]{%
    \multirow{4}{*}{\rotatebox[origin=c]{90}{%
        \begin{tabular}{@{}c@{}}\textbf{#1}\end{tabular}}
        }}

\begin{table*}[t]\centering
\vspace{-.0cm}
\resizebox{.9\textwidth}{!}{
\begin{tabular}{lrrrrrrrrrrrrrrrr}\toprule
& &\multicolumn{3}{c}{\textbf{Number of Q+A pairs}} & &\multicolumn{3}{c}{\textbf{Number of Categories}} & &\multicolumn{2}{c}{\textbf{\% Unique C}} & &\multicolumn{3}{c}{\textbf{Total (I, Q, A) triplets}} \\\cmidrule{3-5}\cmidrule{7-9}\cmidrule{11-12}\cmidrule{14-16}
& &\textbf{Train} &\textbf{Val} &\textbf{Test} & &\textbf{Train} &\textbf{Val} &\textbf{Test} & &\textbf{Val} &\textbf{Test} & &\textbf{Train} &\textbf{Val} &\textbf{Test} \\\midrule

\datasetname{iNat21}     &Templated      &          8,133 &            200 &            500 &&          1,983 &            114 &            300 &&           26\% &           21\% &&         40,665 &          1,000 &          500 \\
                         &Automatic      &         96,317 &          1,000 &          1,500 &&          4,061 &            343 &            562 &&            5\% &            6\% &&        481,585 &          5,000 &        1,500 \\
                         &Multi Answer   &         11,262 &            200 &            500 &&          3,585 &            118 &            318 &&           31\% &           22\% &&         56,310 &          1,000 &          500 \\
                         &Two Hop        &         10,819 &            200 &            500 &&          2,499 &            131 &            359 &&           11\% &            6\% &&         54,095 &          1,000 &          500 \\[1mm]\midrule
\datasetname{GLDv2}      &Templated      &          5,795 &            200 &            500 &&          1,808 &            135 &            268 &&           27\% &           28\% &&         25,870 &            827 &          500 \\
                         &Automatic      &         57,124 &            750 &          1,250 &&          1,965 &            138 &            208 &&            9\% &            9\% &&        255,529 &          3,025 &        1,250 \\
                         &Multi Answer   &         12,667 &            200 &            500 &&          6,449 &            151 &            339 &&           45\% &           56\% &&         56,426 &            844 &          500 \\
                         &Two Hop        &         10,221 &            200 &            500 &&          1,405 &             99 &            154 &&           11\% &           11\% &&         45,771 &            895 &          500 \\[1mm]\midrule
\multicolumn{2}{c}{\textbf{Total}}       &\textbf{212,338}&\textbf{  2,950}&\textbf{  5,750}&&\textbf{ 16,249}&\textbf{  1,071}&\textbf{  2,152}&&\textbf{   14\%}&\textbf{   17\%}&&\textbf{1,016,251}&\textbf{ 13,591}&\textbf{ 5,750}\\
\bottomrule
\end{tabular}
\vspace{-.2cm}
}
\caption{\small \textbf{Dataset Statistics.} We report the number of question, answer, categories and triplets for both supporting datasets and different question types. In total our dataset contains 1M VQA triplets, making it the largest of its kind.
}\label{tab:dataset_statistics_appendix}
\end{table*}

%% file: A5_prompts.tex
\newcommand{\prompt}[3]{%
    \begin{promptfloat}[label={#1}]{#2}
    {\footnotesize \texttt{#3}}
\end{promptfloat}
}

\section{Prompts for dataset creation}

As discussed in Section 4 of the paper, we used a FLAN \cite{chung2022scaling} version of the PaLM \cite{chowdhery22arxiv} model to (1) rephrase automatically generated single-hop questions (both single-answer and multi-answer), and to (2) combine two single-hop questions into a two-hop question.
Generally, we found that the model tends to work better when fed with very specific (and sometimes redundant) instructions, combined with chain-of-thought prompts \cite{wei2022neurips} where possible.
Additionally, we added difficult cases as examples in the prompt, to guide the model's behavior in more challenging situations.

\subsection{Rephrasing automatically generated single-hop questions}
We replaced the name of the category $C$ in the question by its super category.
This rephrasing process was run for all automatically generated single-hop questions (single and multi-answer).
For example, the question \emph{Where is the Église Saint-Cannat located?} was rephrased to \emph{Where is this church located?}
After several iterations, we decided to add three examples in the prompt, using chain-of-thought reasoning with three steps.
The model generally writes the three required steps in the output, and we parse the rephrased question right after the third step written by the model.
Sometimes, the model would return an incorrectly rephrased question.
We observe that the main failure case is when the model simply copies the input question to the output.
Thus, we filter out any rephrased question that is identical to the original question fed in its input.
We sampled $100$ rephrased questions to manually assess the quality of the rephrasings, which we found to be correct in $90\%$ of the cases.
The final prompt is given in Prompt~\ref{prompt:rephrase_single_hop}, where ``\$C\$'' and ``\$Q\$'' are placeholders for the category $C$ and the textual (pre-rephrasing) question $Q$.

\subsection{Chaining single-hop questions into two-hop questions}
We created two-hop questions by chaining two single-hop questions, where the answer to the first single-hop question serves as a bridge entity that makes a connection to the second single-hop question.
For example, given the first single-hop question (SQ1) \emph{What is the main competitor for food for this animal?} with answer (SA1) \emph{Spotted hyena}, and the second single-hop question (SQ2) \emph{What is the population size of this animal?} with answer (SA2) \emph{Between 27,000 and 47,000 individuals}, we generated \emph{What is the population size of the main competitor for food of this animal?}
In this example, the entity \emph{spotted hyena} serves as a bridge between the two single-hop questions.
Note that the answer to the two-hop question is identical to the answer to the second single-hop question, \emph{Between 27,000 and 47,000 individuals}.

After several iterations, we crafted a prompt with 4 examples of varying difficulty (Prompt~\ref{prompt:chaining_two_hop}).
The output of the model is taken as the chained two-hop question.
Sometimes, though, the model would return an incorrect two-hop question.
For this reason, we designed a second prompt to validate the two-hop question provided as the initial output (Prompt~\ref{prompt:validating_two_hop}).
The model is asked to answer the two-hop question using the two initial single-hop questions with answers as context.
If the predicted answer is identical to the answer of the second single-hop question, then the two-hop question is validated and kept in our dataset; otherwise, it is discarded.
Finally, we filter some common failure cases, such as when the model outputs the exact first single-hop or second single-hop question as the chained two-hop question.
We sampled $100$ chained two-hop questions to manually assess their quality, which we found to be correct in $88\%$ of the cases.
In the final prompts (Prompts~\ref{prompt:chaining_two_hop}-\ref{prompt:validating_two_hop}) ``\$SQ1\$'' / ``\$SA1\$'' are placeholders for the first single-hop question / answer, ``\$SQ2\$'' / ``\$SA2\$'' are placeholders for the second single-hop question / answer, and ``\$Q\$'' is the placeholder for the generated two-hop question.

\section{Prompts for evaluation}
In Section 5 we analyze the performance of large models in our dataset.
Similarly to the dataset creation process, we use diverse prompts to adapt and improve the behavior of these models.
In particular, we use prompts to incorporate retrieval results and to identify relevant sections for articles retrieved with Lens.

\subsection{Prompts for question answering}
We use textual Prompts~\ref{prompt:evaluation_question_only}-\ref{prompt:evaluation_oracle_kb_section} to produce answers with large models.
Note that we use the same prompts when using PaLI or PaLM.
However, we always use the question image $\mathcal{I}$ as an additional input to the model when using PaLI.
Note that we use \$Q\$ to refer to the textual part of a question, \$C\$ to denote the question subject $\mathcal{C}$ name, \$Art\$ to denote the full text of a Wikipedia article and \$S\$ to denote a section of a Wikipedia article.
We use the same retrieval prompts for Lens and Oracle setups.
However, for Lens experiments, we only include \$Art\$ or \$S\$ if available, as we might not retrieve entities with Lens or find relevant sections with PaLM.
When not available, we revert to using Prompt~\ref{prompt:evaluation_question_only}.

\subsection{Prompt for relevant section identification}
We use Prompt~\ref{prompt:evaluation_lens_get_kb_section} to identify relevant sections in a Wikipedia article retrieved by Lens.
To do so, we query PaLM for each section \$S\$ in the Lens retrieved Wikipedia article for a question \$Q\$.
Note that we use a few examples in the prompt to condition PaLM to generate a yes/no answer.
The answer produced by PaLM is converted into a string and matched to either \texttt{yes} or \texttt{no}.
When no match is found, we assume that section is not relevant to the question.
If more than one section is identified as relevant for a question, the input \$S\$ to Prompt~\ref{prompt:evaluation_oracle_kb_section} becomes the concatenation of all relevant sections.

\onecolumn

\prompt{prompt:rephrase_single_hop}{Rephrasing automatically-generated single hop question}{In this task, please rephrase the question by replacing the entity name by the word "this", followed by the type of the entity. See the examples below:\\
\\
EXAMPLE 1:\\
  entity name: eiffel tower\\
  question: How tall is the eiffel tower?\\
  step 1 (find type of entity): The eiffel tower is a type of: tower\\
  step 2 (write the word "this" followed by the type obtained in "step 1"): this tower\\
  step 3 (final rephrased question): How tall is this tower?\\
  \\
  EXAMPLE 2:\\
  entity name: salmon\\
  question: Which country is the largest producer of salmon?\\
  step 1 (find type of entity): The salmon is a type of: fish\\
  step 2 (write the word "this" followed by the type obtained in "step 1"): this fish\\
  step 3 (final rephrased question): Which country is the largest producer of this fish?\\
  \\
  EXAMPLE 3:\\
  entity name: Grand Beach Provincial Park\\
  question: grand beach provincial park is located on the east side of what lake?\\
  step 1 (find type of entity): The Grand Beach Provincial Park is a type of: park\\
  step 2 (write the word "this" followed by the type obtained in "step 1"): this park\\
  step 3 (final rephrased question): this park is located on the east side of what lake?\\
  \\
  Note that the entity name should not be part of the rephrased question.\\
  Please make sure to write out all 3 steps as in the examples above.\\
  \\
  Based on the above examples, provide a rephrased question for the following case:\\
  entity name: \$C\$\\
  question: \$Q\$\\}

\prompt{prompt:chaining_two_hop}{Chaining two single-hop questions into a two-hop question}{EXAMPLE 1:\\
  question 1: in which city is this building located?\\
  answer 1: San Francisco\\
  question 2: what is the average temperature in San Francisco?\\
  answer 2: 15 Celsius\\
  combined question: What is the average temperature in the city where this building is located?\\
\\
  EXAMPLE 2:\\
  question 1: what is the predator of this animal?\\
  answer 1: Lion\\
  question 2: What is the weight of a lion on average?\\
  answer 2: 190 kilograms\\
  combined question: What is the average weight of the predator of this animal?\\
\\
  EXAMPLE 3:\\
  question 1: In what country is this plant found?\\
  answer 1: Australia\\
  question 2: What does australia's size give it a wide variety of?\\
  answer 2: landscapes and climates\\
  combined question: What does the size of the country where this plant is found give it a wide variety of?\\
\\
  EXAMPLE 4:\\
  question 1: What country is this plant the national flower of?\\
  answer 1: South Africa\\
  question 2: south africa is a member of the commonwealth of nations and what other organization?\\
  answer 2: the G20\\
  combined question: The country that this plant is the national flower of is a member of the commonwealth of nations and what other organization?\\
\\
  Based on the above 4 examples, provide a combined question for the following case, such that the answer to the combined question is the same as the answer to question 2:
\\
  question 1: \$SQ1\$\\
  answer 1: \$SA1\$\\
  question 2: \$SQ2\$\\
  answer 2: \$SA2\$\\
  combined question:}

\prompt{prompt:validating_two_hop}{Validating two-hop question given the original single-hop questions}{question 1: \$SQ1\$\\
  answer 1: \$SA1\$\\
  question 2: \$SQ2\$\\
  answer 2: \$SA2\$\\
  Based on the questions and answers above, please answer the following question: \$Q\$
}

\prompt{prompt:evaluation_question_only}{Question only evaluation}{Question: \$Q\$\\The answer is: }
\prompt{prompt:evaluation_oracle_entityname}{Retrieval with entity name}{Entity name: \$C\$\\Question: \$Q\$\\The answer is: }
\prompt{prompt:evaluation_oracle_kb_article}{Retrieval with KB Article}{Context: \$Art\$\\Question: \$Q\$\\The answer is: }
\prompt{prompt:evaluation_oracle_kb_section}{Retrieval with KB Section}{Context: \$S\$\\Question: \$Q\$\\The answer is: }
\prompt{prompt:evaluation_lens_get_kb_section}{Lens retrieval to get KB Section}{%
Can the answer to the question be found in the text?\\
Question: Is this fungus edible?\\
Text: This compound induces mammalian cells (specifically, the cell line HL60 to differentiate into granulocyte- or macrophage-like cells. The fungus also contains the mycotoxin muscarine, and the antifungal metabolite strobilurin D. Despite the presence of these toxins, some guides list this fungus safe for human consumption.\\
The answer is: yes\\
\\
Can the answer to the question be found in the text?\\
Question: In which season does this plant give flowers?  \\
Text: This cactus has stems about 1/2-1 inch wide with 6-9 edges. Its flowers are white, up to 30 centimetres in diameter with a scent redolent of vanilla. The flowers open after sundown, closing and wasting after a few hours. By 9 am the next day they are gone. \\
The answer is: no \\
\\
Can the answer to the question be found in the text? \\
Question: What is the habitat of this animal? \\
Text: X is native to Europe and North Africa through to Central Asia. It is introduced to the United States and parts of South America. It widespread across the northeastern United States and eastern Canada, and can be found outside, or more commonly inside houses. It is thought to have been introduced into America from Europe by English colonists. \\
The answer is: yes \\
\\
Can the answer to the question be found in the text? \\
Question: \$Q\$ \\
Text: \$S\$ \\
The answer is:
}